\documentclass[smallextended]{svjour3}

\usepackage{times}
\usepackage{float}
\usepackage{graphicx}
\usepackage{epsfig}
\usepackage{setspace}
\usepackage{multirow}
\usepackage{amsmath,amssymb}
\usepackage{amsfonts}
\usepackage{color}
\usepackage{mathptmx}
\usepackage{textcomp}
\usepackage{siunitx}
\usepackage[left=24mm, right=24mm]{geometry}
\usepackage{url}

\begin{document}

\title{Remote Detection of Idling Cars Using Infrared Imaging and Deep Networks}

\author{Muhammet Bastan \and Kim-Hui Yap \and  Lap-Pui Chau}
\institute{
Muhammet Bastan (corresponding author) \at
School of Electrical and Electronic Engineering\\
Nanyang Technological University, 639798, Singapore\\
\email{mubastan@gmail.com}
\and
Kim-Hui Yap \at
School of Electrical and Electronic Engineering\\
Nanyang Technological University, 639798, Singapore\\
\email{ekhyap@ntu.edu.sg}
\and 
Lap-Pui Chau \at
School of Electrical and Electronic Engineering\\
Nanyang Technological University, 639798, Singapore\\
\email{elpchau@ntu.edu.sg}
}

\date{Date: April 2018}

\maketitle

\begin{abstract}

Idling vehicles waste energy and pollute the environment through exhaust emission. In some countries,  idling a vehicle for more than a predefined duration is prohibited and automatic idling vehicle detection is desirable for law enforcement. We propose the first automatic system to detect idling cars, using infrared (IR) imaging and deep networks.

We rely on the differences in spatio-temporal heat signatures of idling and stopped cars and monitor the car temperature with a long-wavelength IR camera. We formulate the idling car detection problem as spatio-temporal event detection in IR image sequences and employ deep networks for spatio-temporal modeling.
We collected the first IR image sequence dataset for idling car detection.
First, we detect the cars in each IR image using a convolutional neural network, which is pre-trained on regular RGB images and fine-tuned on IR images for higher accuracy. Then, we track the detected cars over time to identify the cars that are parked. Finally, we use the 3D spatio-temporal IR image volume of each parked car as input to convolutional and recurrent networks to classify them as idling or not. We carried out an extensive empirical evaluation of temporal and spatio-temporal modeling approaches with various convolutional and recurrent architectures. We present promising experimental results on our IR image sequence dataset.

\keywords{infrared image \and car detection \and idle detection \and deep neural networks}

\end{abstract}

\section{Introduction}

Fuel consumption and exhaust emissions are higher for idling over 10 seconds, compared to restarting~\cite{idle-2012}. Idling for 6 minutes a day costs the drivers a total of $10$ billion dollars a year in the USA~\cite{idle-2012}. Increased exhaust emission is another unpleasant consequence of idling.

The Paris Climate Agreement~\cite{pca-ajil16}, adopted in 2015, aims to mitigate greenhouse gas emissions and reduce global warming and air pollution. Reducing vehicular emission is important to contribute to the agreement. One way to reduce vehicular emission is to reduce idling time. In line with this, in Singapore, it is prohibited to keep a vehicle idling for more than $3$ minutes, for reasons other than traffic conditions. Currently, this regulation is enforced by human officers by manually checking if a parked vehicle is idling or not.

It is desirable to employ an automatic system for monitoring, since manual checking is not practical. Moreover, the automatic system can also archive evidence of the infringement. Motivated by this, we propose the first automated system to detect idling cars by monitoring the car temperature with the help of a thermal infrared (IR) camera. The intuition is that there should be differences between the spatio-temporal heat signatures of idling and stopped cars. In an idling car, the engine is running, producing and dissipating heat; the fan cooling is active to keep the engine temperature at a specific level; the air conditioning may be active to regulate the internal car temperature. When the engine is turned off, the heat inside the engine starts to dissipate out, first heating up the car for some time, and then cooling down; the cooling system and air conditioning are inactive. These all contribute to the spatio-temporal heat signature of a car, which can be monitored remotely with an IR camera.

\begin{figure}[ht]    
    \centerline{\includegraphics[width=1.0\textwidth]{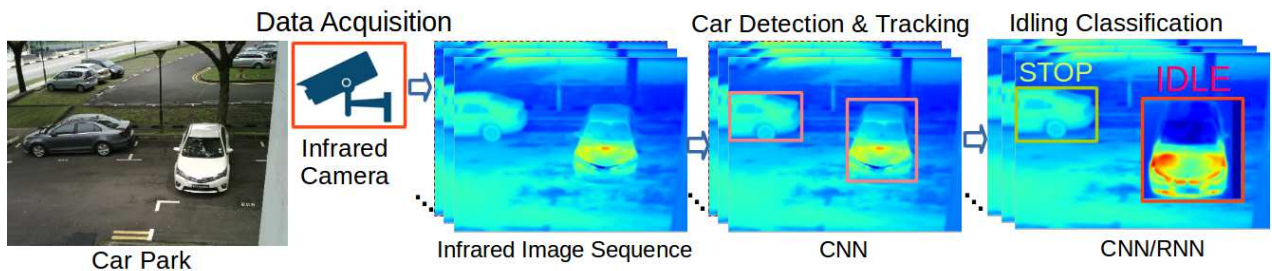}} 
    \caption{Proposed idling car detection framework: (1) Data acquisition, (2) Car detection with convolutional neural networks (Faster R-CNN) and tracking, (3) Idling classification using convolutional and recurrent networks.}
    \label{fig:system}
\end{figure}

To the best of our knowledge, there is no published work that can remotely detect idling cars, using thermal infrared cameras or other sensors. This work is the first to collect a dataset and propose an automated remote idling car detection system using an IR camera. This paper is an extended version of \cite{idle-iscas18}, which presents an idling car detection framework using convolutional neural networks.
This paper additionally includes (i) evaluation of temporal modeling approaches for idling detection, (ii) evaluation of spatio-temporal modeling for idling detection using convolutional and recurrent networks, (iii) more detailed description and analysis of the IR car image sequence dataset, and (iv) more detailed description and analysis of car detection and idling classification using convolutional neural networks.

\subsection{Infrared Imaging}

Thermal infrared cameras are passive sensors that capture the infrared radiation emitted by objects and form an image in which the intensities correspond to temperature values~\cite{ir-book-2017,thermal-mva2014}. They operate in the infrared portion of the electromagnetic spectrum. They were originally designed for military applications~\cite{ir-paa14}, surveillance and night vision. Currently, they are being used for more diverse applications, such as gas leakage detection, building heat loss inspection, object detection~\cite{ir-car-prl2006,od-paa17}, pedestrian detection~\cite{pedestrian-ir,pedir-nca16}, face detection and recognition~\cite{faceprint,face-ir,face-ir-paa14}, and health applications~\cite{heart-ir} owing to their falling prices. They are still expensive compared to RGB cameras.

IR cameras measure the surface temperature of the objects, which depends on the emissivity of the surfaces, which in turn depends on the viewing angle~\cite{ir-book-2017}. Contrary to common misconception, thermal infrared cameras cannot see through walls, objects, or even glass. In our case, we can only observe the surface temperature of a car, and not the actual temperature inside the engine. The temperature inside the engine is around $600-700$ \textdegree{}C, while the operating temperature of a car (coolant temperature) is around $90-105$ \textdegree{}C; the surface temperature outside the car (the measured temperature) is below this value. The exhaust temperature can be slightly higher than the operating temperature.

\subsection{Method Overview}

We propose the framework shown in Figure~\ref{fig:system} to detect idling cars. First, we use an IR camera to obtain IR image sequences, hence the temperature, of the whole car park. Then, we apply car detection on each IR image in the sequence to localize the cars and then, further track them to identify stationary/parked cars. Finally, we classify the detected stationary cars as idling or not. We experimented with two different approaches for classification: (1) Using the temporal evolution of maximum car temperature sequence as an input feature to a classifier (Section~\ref{sec:temp-cls}). (2) Modeling the spatio-temporal change of car temperature with convolutional and recurrent neural networks that learn the useful features automatically from the data (Section~\ref{sec:sp-temp-cls}).

\section{Data Collection}
There is no dataset available for idling car detection, therefore, we collected our own dataset, named Car Infrared (CIR) dataset.
We used a long wavelength infrared (LWIR) camera (Testo 885-2) with a spectral range of $7.5$ to $14$ \si{\micro\metre}. The IR image resolution is $320\times240$ pixels, RGB image is 3.1 MP. Available temperature measuring ranges are $-30$ to $+100$ \textdegree{}C, $0$ to $+350$ \textdegree{}C and $0$ to $+650$ \textdegree{}C. The camera can automatically record IR and RGB images at every 5 seconds; higher recording frequency is not supported by this camera. We used only the IR images in this work.

Dataset construction is an expensive process, in our case, comprising both data recording and annotation.
We could install the IR camera at a car park, record all the cars and annotate the recorded sequences. This approach has privacy issues and needs legal permission. Moreover, we might not have collected sufficient samples for the idling cases. It would also be difficult to annotate later by just watching the sequences without knowing whether the car is idling or not.
All in all, we preferred a controlled data acquisition strategy, in which we record cars in a predefined setting, with the help of volunteering car owners.

We setup the camera on a tripod, watching a car park at an altitude of $3-4$ meters, similar to surveillance cameras. The distance between the camera and cars was $5-10$ meters. With this setup, we recorded 8 different cars with brands BMW, Honda, Kia, Mitsubishi (2), Toyota (2) and Volvo, in three views (front, rear, side) for at least $5$ minutes. We recorded $1$ IR and RGB frame every $5$ seconds. To imitate a realistic scenario, the car first drives around for a few minutes, then parks in front/rear/side view and keeps the engine idling for at least 5 minutes, while being recorded. This is repeated for the case of a stopped engine. 

We refer to the recording of each of the parking in a specified view and engine state as a \textit{Sequence S}, e.g., $S_1$: a car is parked in front view, engine idling for 5 minutes; $S_2$: a car is parked in rear view, engine stopped for 5 minutes. Hence, there are at least 6 sequences for each car, one for each view and engine state combination (front+idle, side+idle, rear+idle, front+stop, side+stop, rear+stop). We performed the recordings during daylight and non-rainy weather; the ambient temperature was around $30$ \textdegree{}C (Singapore). We annotated the dataset by bounding boxes around the cars, and the view (front, side, rear) and engine state (idling, stopped) for each car.

\begin{figure}[ht]    
    \centerline{\includegraphics[width=0.8\textwidth]{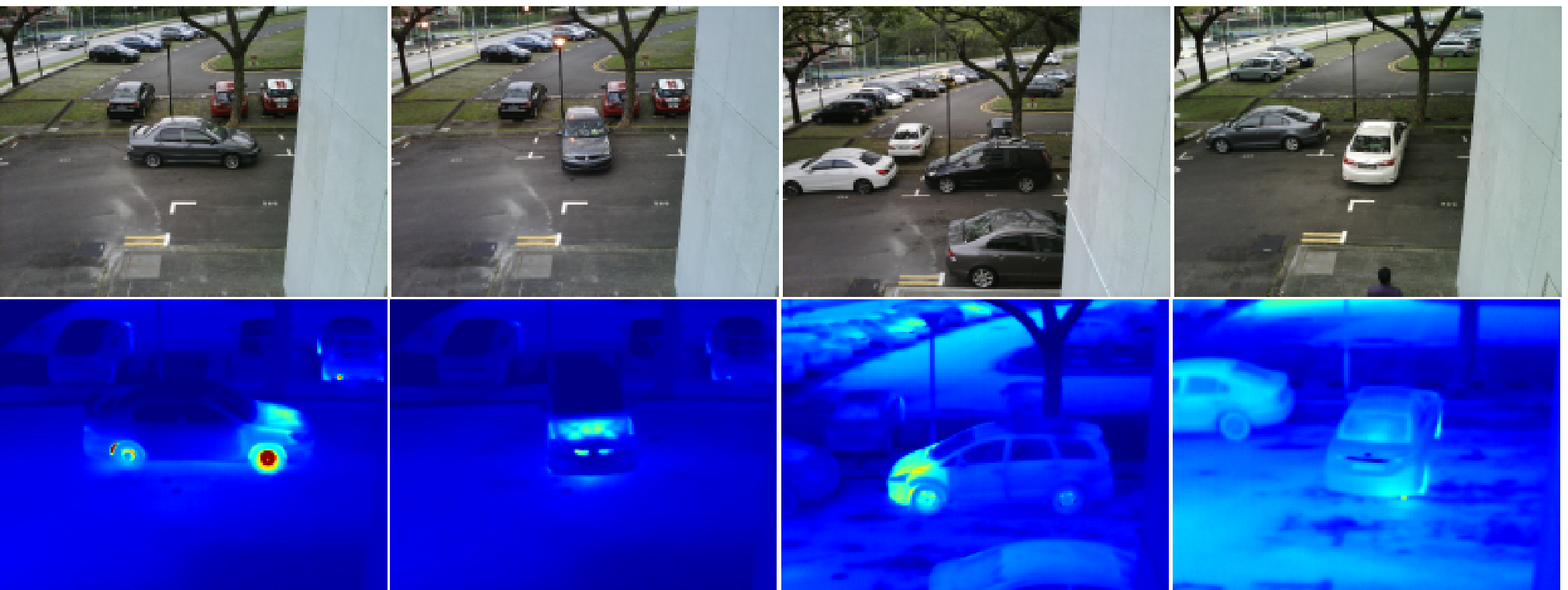}}
    \vspace{1mm}
    \centerline{\includegraphics[width=0.8\textwidth]{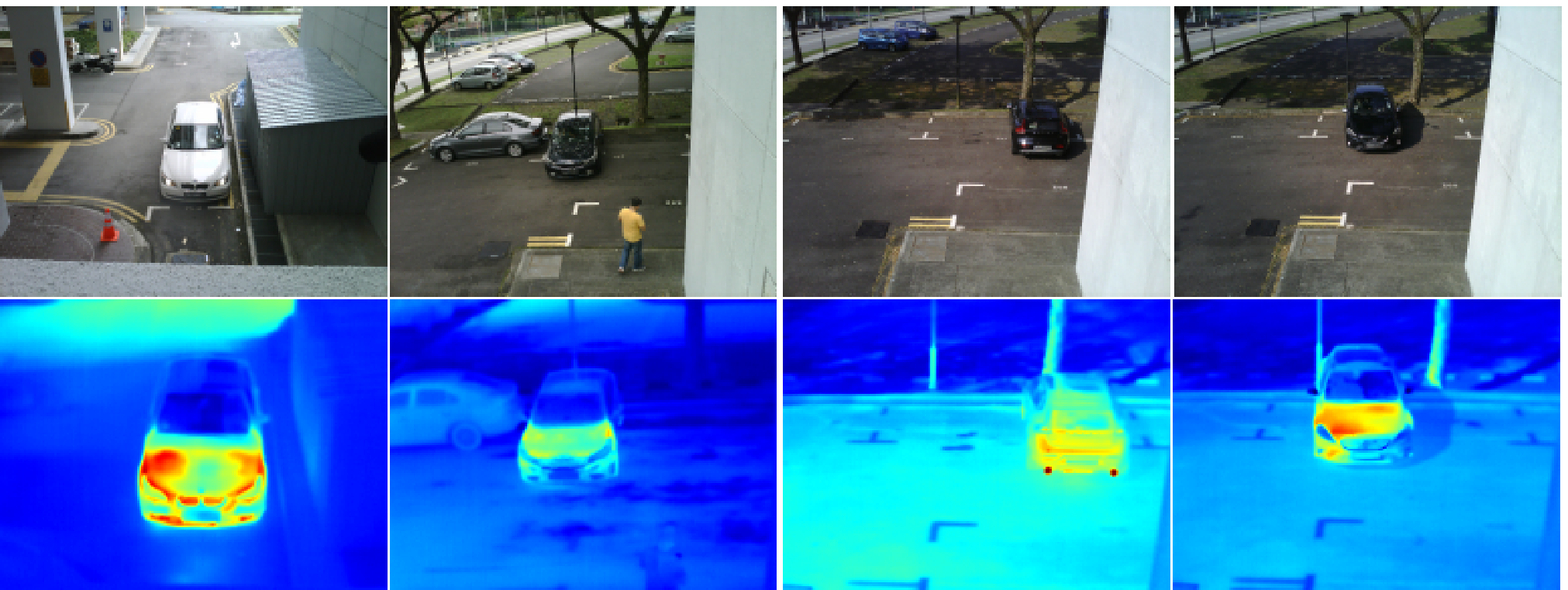}}
    
    \caption{Sample RGB and IR images from the dataset. RGB images have higher resolution and cover a larger area than IR images. In the figures, IR images are converted to RGB heat maps for better visualization.}
    \label{fig:dataset-samples}
\end{figure}

Our dataset contains $5670$ images, $53$ idling and stop sequences, corresponding to about $8$ hours of recording. The IR images are single channel grayscale images; pixel values represent the temperature in \textdegree{}C. The temperature values are decoded using the default camera parameters (emissivity: $0.96$) with the help of the proprietary library provided by the vendor. Although the emissivity values are different for different surfaces, we are more interested in the relative spatio-temporal change of temperature over time, rather than accurate absolute temperature values. Figure~\ref{fig:dataset-samples} shows sample RGB and IR images from the dataset. In all the figures, the single channel IR images are converted to RGB heat maps for better visualization.

\begin{figure}[ht]    
    \centerline{BMW}
    \centerline{\includegraphics[width=1.0\textwidth]{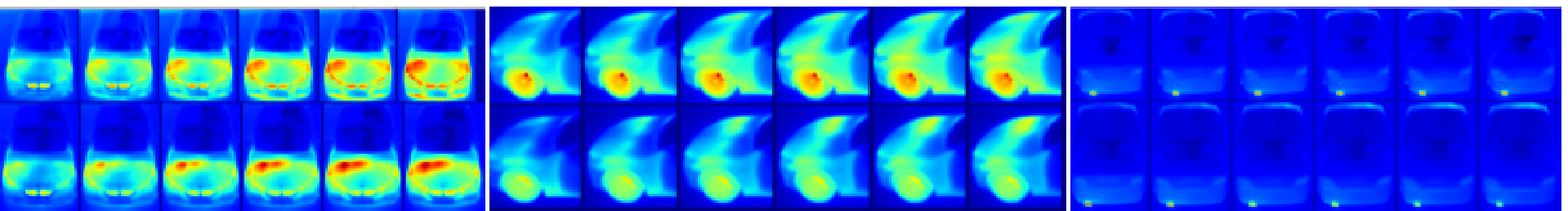}}
    \vspace{1mm}
    \centerline{Volvo}
    \centerline{\includegraphics[width=1.0\textwidth]{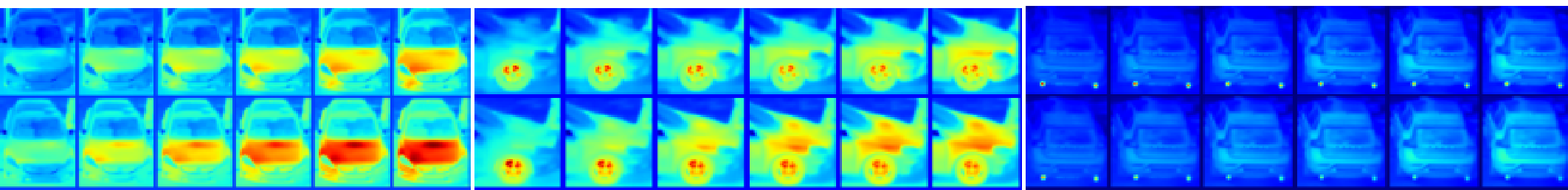}}
    \caption{Front, side and rear view cropped LWIR images of a BMW and a Volvo over 5 minutes, sampled at 1 minute time intervals. Top rows: engine idling, bottom rows: engine stopped. The Volvo got direct sunlight during recording.}
    \label{fig:ir-image-view}
\end{figure}

\begin{figure}[ht]    
    
    \centerline{\includegraphics[width=0.33\textwidth]{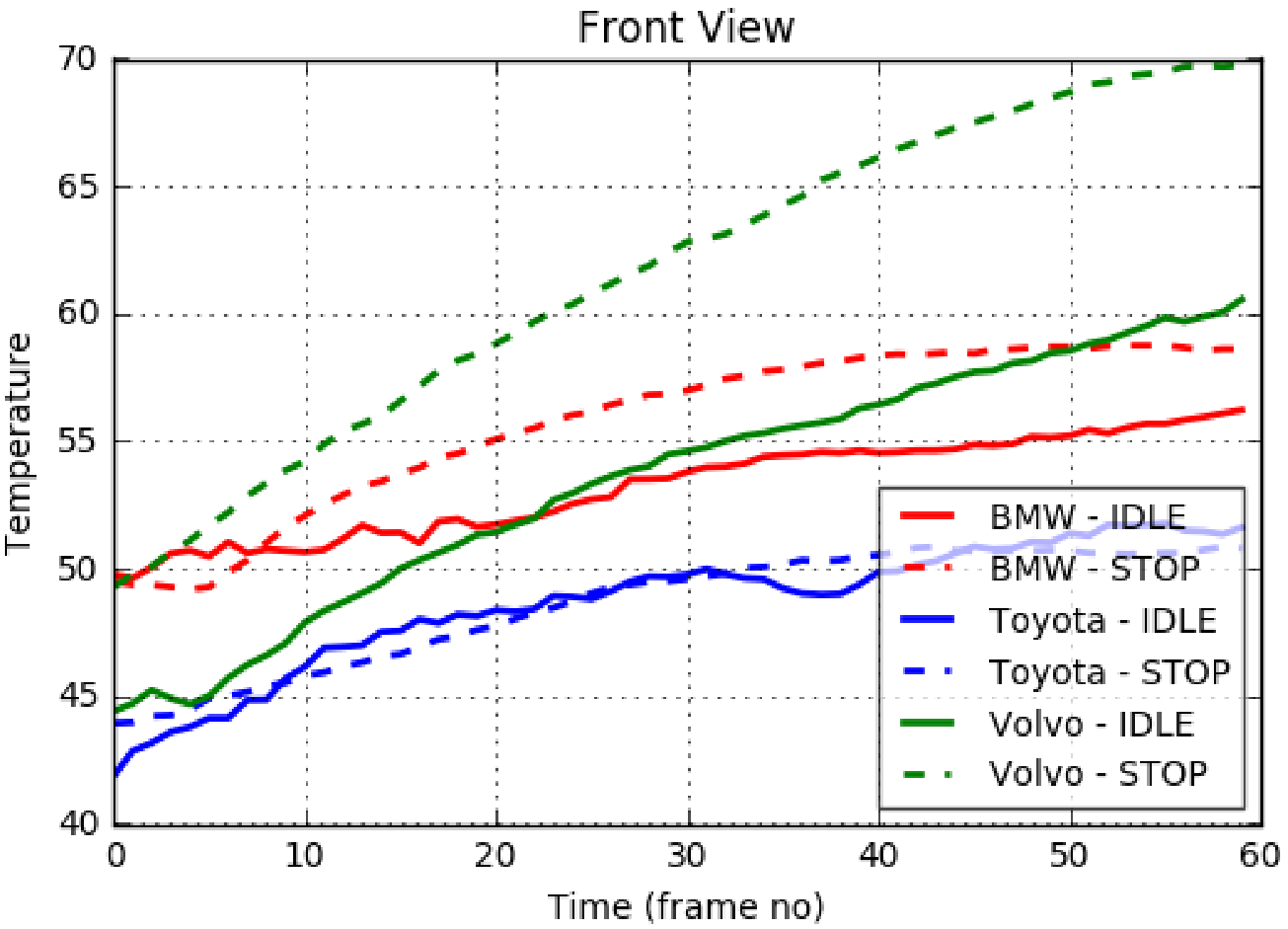}
		\includegraphics[width=0.33\textwidth]{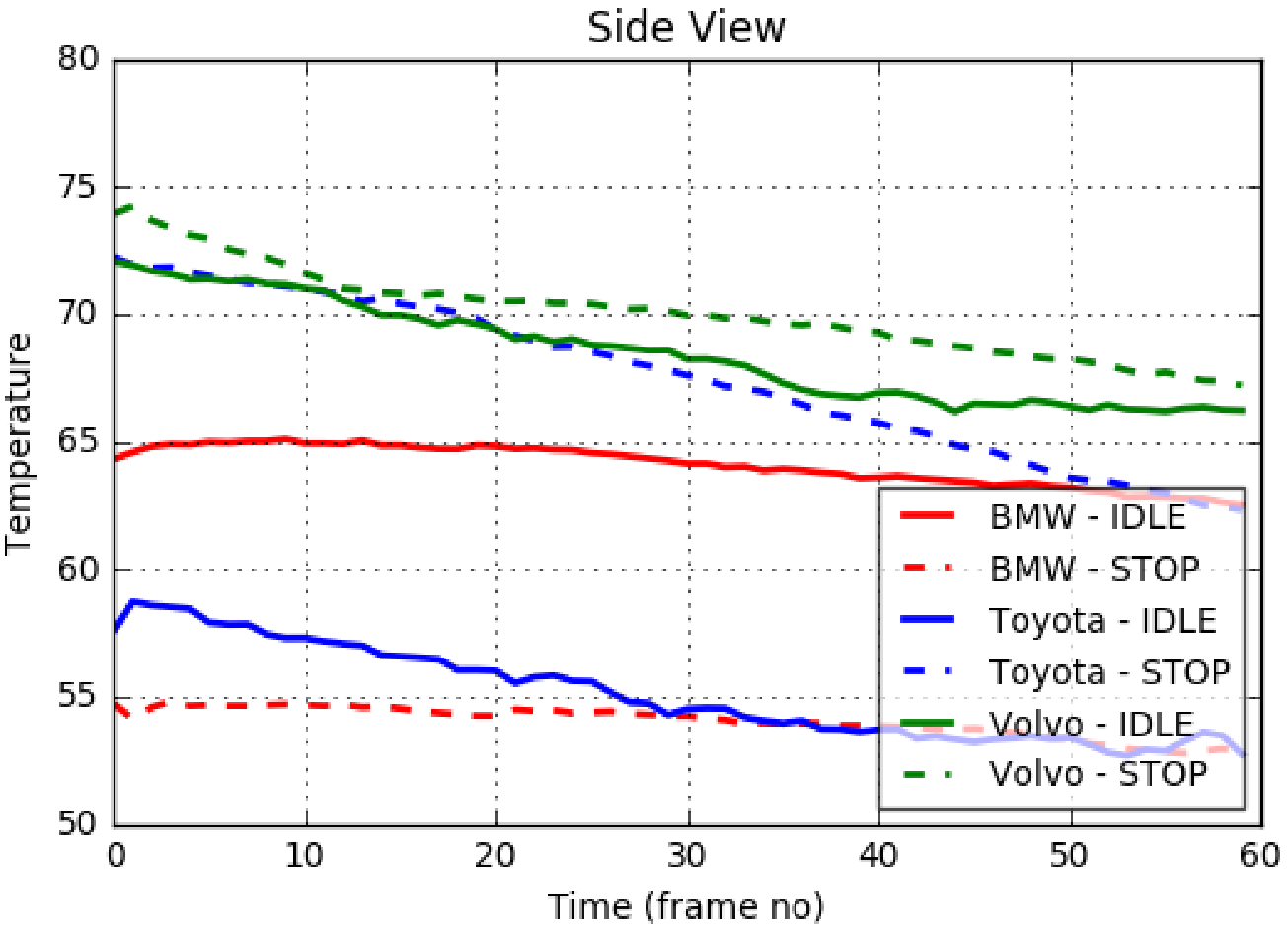}
		\includegraphics[width=0.33\textwidth]{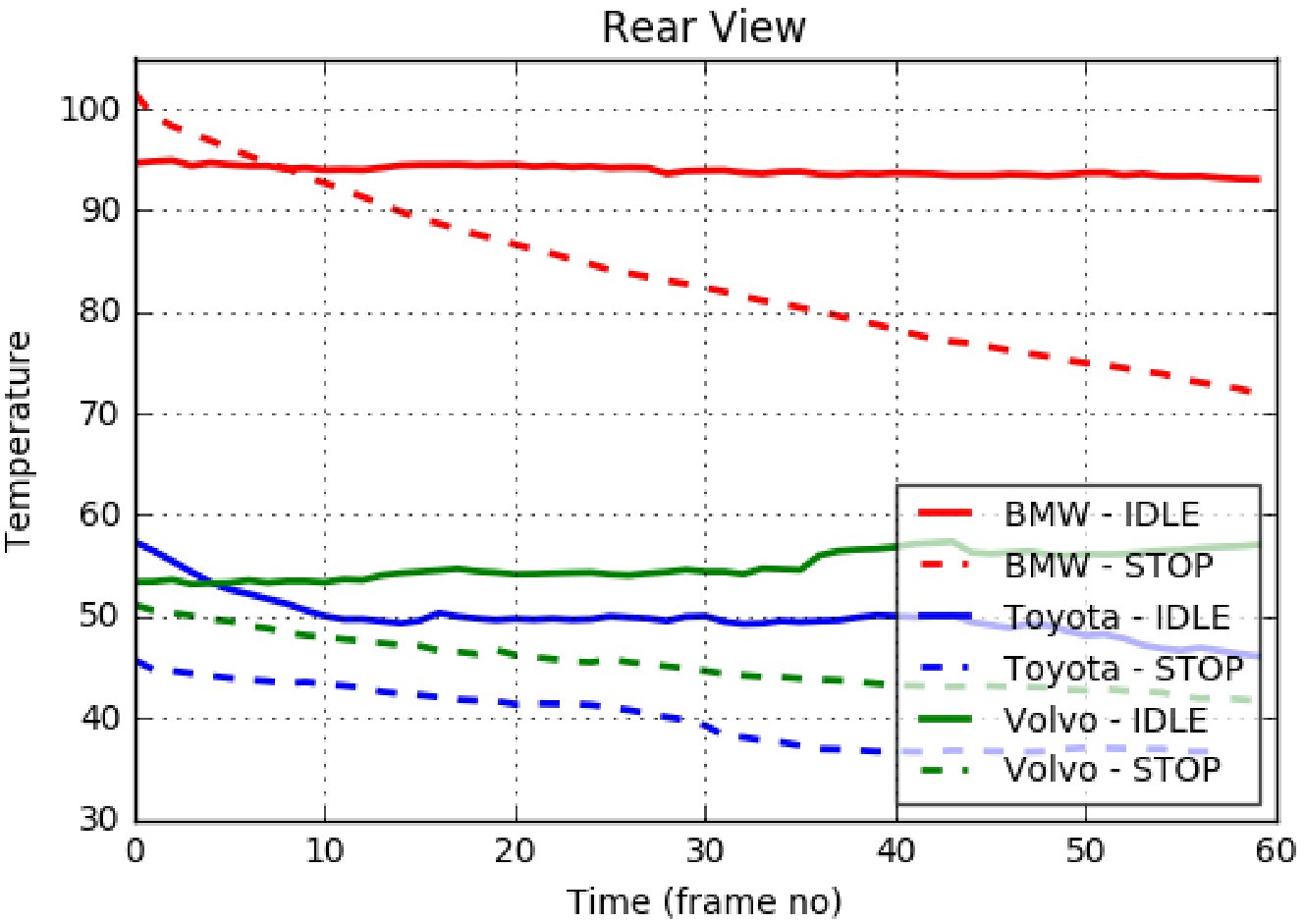}}
    \caption{Temporal evolution of maximum temperatures in front, side and rear views for 3 different cars over 5 minutes (60 frames). The temperatures are in \textdegree{}C.}
    \label{fig:ir-temp-graphs}
\end{figure}

Figure~\ref{fig:ir-image-view} shows the spatio-temporal change of IR image (temperature) of a BMW and Volvo in front, side and rear view over 5 minutes. Figure~\ref{fig:ir-temp-graphs} shows the evolution of maximum temperatures over 5 minutes for different cars. The evolution of the temperature depends on both internal (engine, cooling, air conditioning) and external factors (wind, sun, rain, ambient temperature).
In front view, right after a car is parked, its temperature keeps increasing over 5 minutes, when the engine is idling or stopped. In rear view, the maximum temperature corresponds to the exhaust temperature and it decreases when the engine is stopped; when idling, it increases if the initial temperature is low, e.g., 50 \textdegree{}C; it decreases or stays the same when the initial temperature is high.
In side view, the maximum temperature initially corresponds to the tire break temperature and it decreases for a few minutes, as shown in the graph. The maximum temperature at the hood (not shown in the graphs) evolves similarly to the front view.

Judging by these observations, it may be possible to differentiate idling and stopped cars in rear view by using the maximum temperature profiles as shown in Figure~\ref{fig:ir-temp-graphs}; however, exhaust pipes of some cars are hidden and not visible in the IR images (there is one such car in our dataset). In front view, maximum temperature profiles are hard to differentiate. Fortunately, there are spatio-temporal differences, which are also clearly visible in the given images in Figure~\ref{fig:ir-image-view}. However, spatio-temporal profiles are not common across all types of different cars; they are similar in some cars and somewhat different in others. Overall, the problem of differentiating idling cars from non-idling cars is not trivial even when they are localized correctly.

\section{Car Detection}

The first step in our idling car detection framework (Figure~\ref{fig:system}) is the localization of stationary cars in the image sequence.
The localization can be done in RGB or IR images. However, RGB and IR images are not aligned; they need to be registered to find the corresponding bounding box locations~\cite{irgb-reg-mva18}. Moreover, localization in RGB images will not work well under low illumination, e.g., at night or in dark closed car parks. Therefore, localization in IR images is preferable. Our work is the first to report car detection performance on IR images.

We employed a convolutional network based car detector, namely Faster R-CNN~\cite{faster-rcnn}, to localize the cars in IR images. 
There are other more efficient/faster alternative CNN-based object detectors, such SqueezeDet~\cite{squeezedet}, SSD~\cite{ssd-eccv2016}, YOLO~\cite{yolo} and YOLO9000~\cite{yolo9000}, however, the localization accuracy of Faster R-CNN is usually better.

Deep networks require abundant labeled data for training. Transfer learning~\cite{transfer-cvpr15,pedir-nca16} is one way to mitigate the small dataset problem. However, the publicly available pre-trained networks are usually trained on standard RGB image datasets, like ImageNet~\cite{imagenet-ijcv2015} or MS COCO~\cite{ms-coco2014}. IR images are single channel and not compatible as input to those pre-trained networks. There are two ways to overcome this: (1) convert single channel IR image to 3 channels by duplicating the single channel and use standard pre-trained networks, (2) train a network with single channel input on grayscale images of large datasets, then fine-tune on IR images. The second approach is costly, especially when the network is large, which needs larger training datasets. We tried both approaches and found that pre-training on regular RGB/grayscale images, and then fine tuning on IR images improve the detection performance.

We adapted the publicly available Faster R-CNN implementation at \url{https://github.com/yhenon/keras-frcnn}, using Keras~\cite{keras} library with TensorFlow~\cite{tensorflow} backend. Based on our dataset, we changed the minimum image size to $480$ (doubling the IR image size), anchor box scales to $[200, 350, 500]$, anchor box ratios to $[(1, 1), (1, 1.5), (2, 1)]$, and used horizontal flip as the only data augmentation.

\begin{figure}[ht]    
    \centerline{\includegraphics[width=0.45\textwidth]{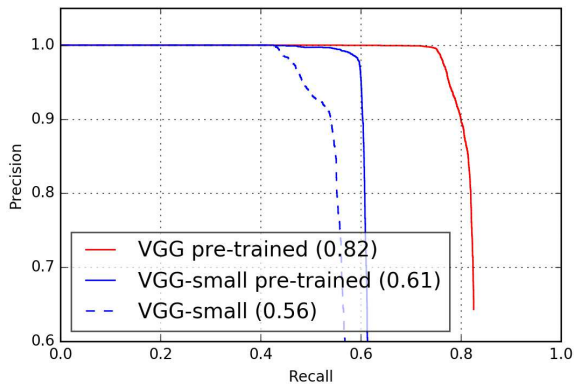}} 
    \caption{Car detection performance with Faster R-CNN and VGG base networks. `VGG' uses 3 channel images as input, while `VGG-small' uses single channel images. Average precision values are shown inside parenthesis. Pre-training on regular RGB/grayscale images improves detection performance.}
    \label{fig:detection-per}
\end{figure}

We divided the dataset into two parts; trained on the first part and tested on the second part and vice versa, and combined the results to evaluate the performance (i.e., 2-fold cross validation).
Figure~\ref{fig:detection-per} shows the car detection performance on our dataset. For transfer learning, we used the standard procedure: first trained the last classification layer, freezing all the previous layers, then trained the whole network. For the VGG base network~\cite{vgg-iclr15}, we use an ImageNet~\cite{imagenet-ijcv2015} pre-trained network, fine tune it on PASCAL VOC~\cite{pascal-voc2010} car detection, and finally fine tune it on IR car detection. `VGG-small' is a much smaller network with 1/10 parameters of original VGG, it was trained directly on the single channel IR images. `VGG-small pre-trained' network is first pre-trained on grayscale PASCAL VOC for car detection, and then fined tuned on IR car detection; this improved average precision by 5 points, as shown in the graph. 

\begin{figure}[ht]    
    \centerline{\includegraphics[width=0.9\textwidth]{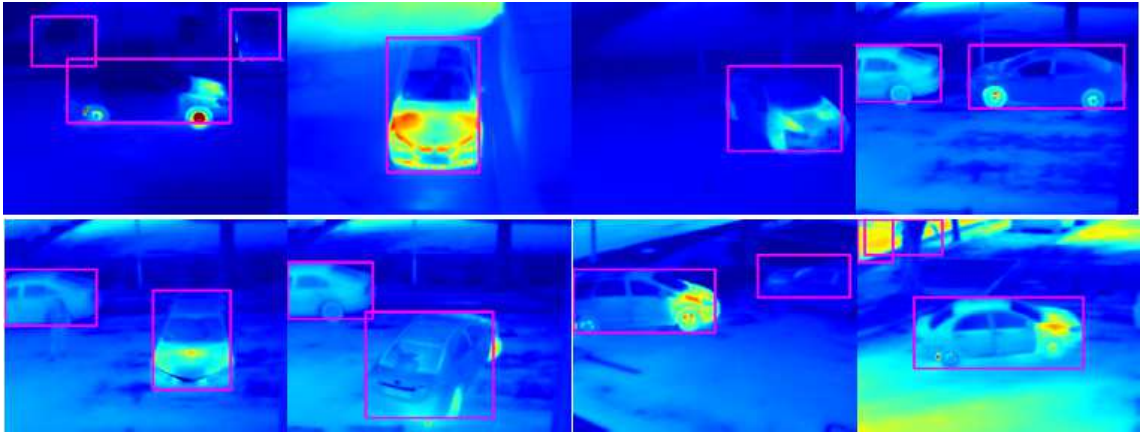}
    } 
    \caption{Car detection examples on IR images with Faster R-CNN, VGG base network, using 3-channel input images and pre-trained on PASCAL VOC 2007 RGB images, fine-tuned on IR images.}
    \label{fig:detection-samples}
\end{figure}

Figure~\ref{fig:detection-samples} shows example detections with the VGG base network. The detector with VGG base network works fairly well, especially when the car temperature is higher than the ambient temperature; this is the case we are interested in. The localization is not perfect, but this is also the case in RGB images. The car detection performance on our IR dataset is in agreement with the car detection performance of 0.79 average precision on PASCAL VOC using RGB images. Moreover, we would obtain the same detection performance on IR dataset even if the dataset were recorded at night, in which case detection on RGB images would deteriorate significantly.

Next step after car detection is the determination of the cars that are parked/stationary for at least 3 minutes (36 frames). We do so by using a simple tracking algorithm based on the car bounding boxes, which should not change much if the car is stationary. If the intersection over union between bounding boxes of consecutive frames is higher than $0.6$, the boxes are assigned to the same car track. Tracks shorter than 3 minutes, and those with average detection score below $0.9$ are discarded. This also eliminates some of the false detections.
Finally, the bounding boxes belonging to the same car track are averaged to reduce the affect of fluctuations in the bounding boxes throughout the sequence. The bounding box is fixed, since the car is stationary. This average bounding box is used in the subsequence idling classification.

\section{Temporal Modeling for Idling Classification}
\label{sec:temp-cls}

Temporal idling classification uses the temporal evolution of maximum car temperature as the input feature. Figure~\ref{fig:ir-temp-graphs} shows examples of how maximum temperature changes over time for different cars in front view, rear view and side view. Each recorded sequence is at least $5$ minutes ($60$ frames). We take subsequences of length $3$ minutes ($36$ frames) to increase the number of training and test samples. For each sequence $S_i$, a subsequence $S_{ij}$ is a subsequence of length $36$ frames, starting at frame $j$. Hence, each feature vector is $36$ dimensional, corresponding to the maximum temperatures over the car bounding box for $3$ minutes ($36$ frames). We shifted each subsequence by subtracting the first value, so that each subsequence starts at zero. This is to reduce the affect of different initial temperatures. We use a maximum of $30$ subsequences for each sequence.

This is intrinsically a sequence modeling problem, therefore, sequence modeling methods should work better. We experimented with several classifiers and reported the results for the following.

\begin{itemize}
 \item Support Vector Machines (SVMs) with RBF kernel, $C=0.5$, with probability estimates. We used the implementation in Scikit-learn library~\cite{scikit-learn}. SVMs are not specifically good for sequence modeling.
 
 \item 1D convolutional neural network (CNN) with $3\times1$ convolution $C(3,1)$, which is able to model the temporal sequence data. The network structure is 64$C(3,1)$, 64$C(3,1)$, MaxPool(2), 128$C(3,1)$, 128$C(3,1)$, MaxPool(2), Dropout(0.5), FC(128), Dropout(0.5), FC(2), Softmax, with \textit{relu} activations.
 
 \item Long-short term memory (LSTM) network~\cite{lstm-1997}, which is known for its ability to capture long range dependencies in sequence data. The network structure is LSTM(512, Dropout=0.5), FC(128), Dropout(0.5), FC(2), Softmax.
\end{itemize}

We used the Adam optimizer~\cite{adam-2014} with a learning rate of $0.0001$ to minimize categorical cross entropy loss in both 1D CNN and LSTM networks. We used TFLearn (TensorFlow)~\cite{tflearn} for implementation.

\begin{figure}[ht]    
    \centerline{\includegraphics[width=0.33\textwidth]{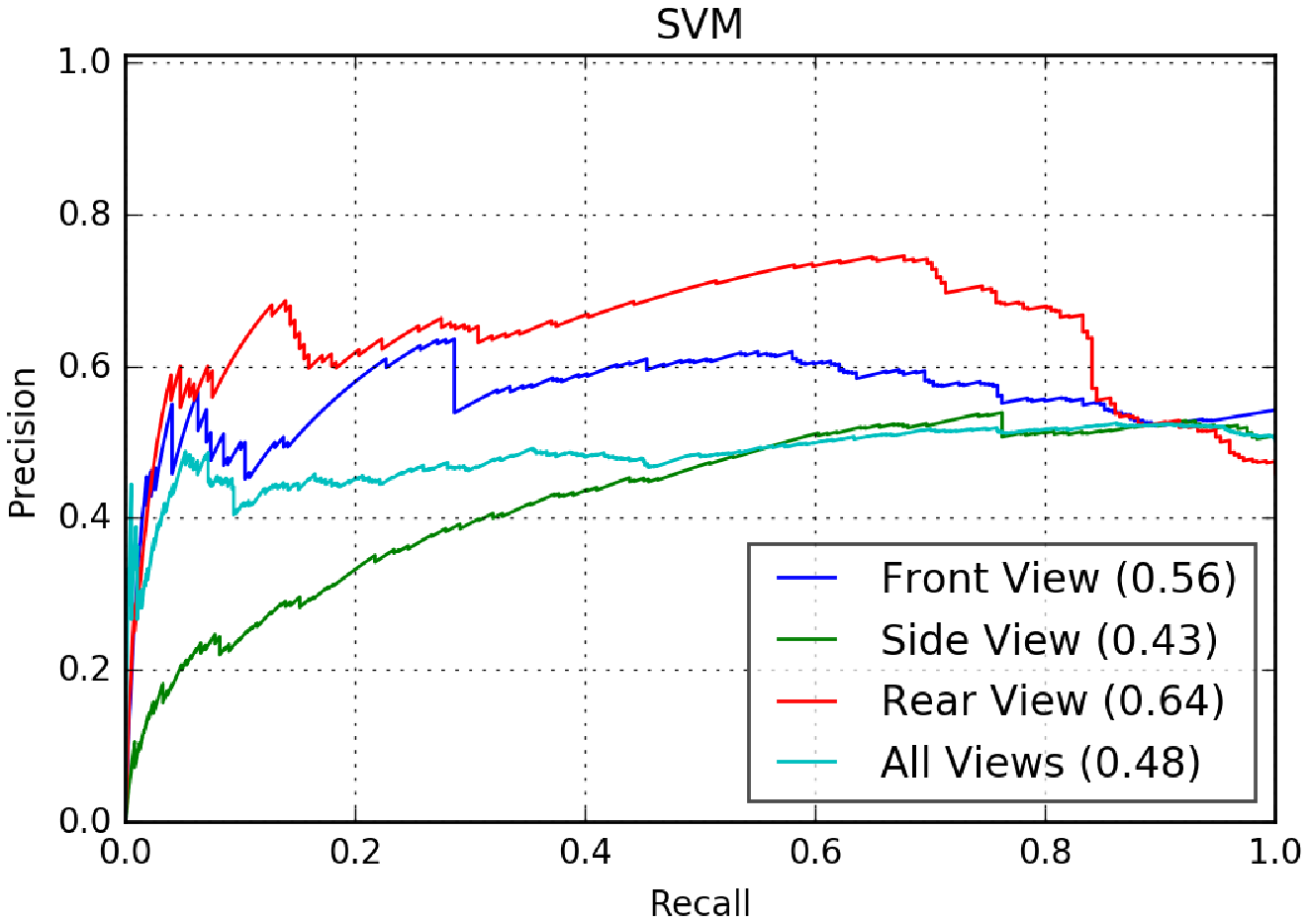}
    \hspace{1mm}
    \includegraphics[width=0.33\textwidth]{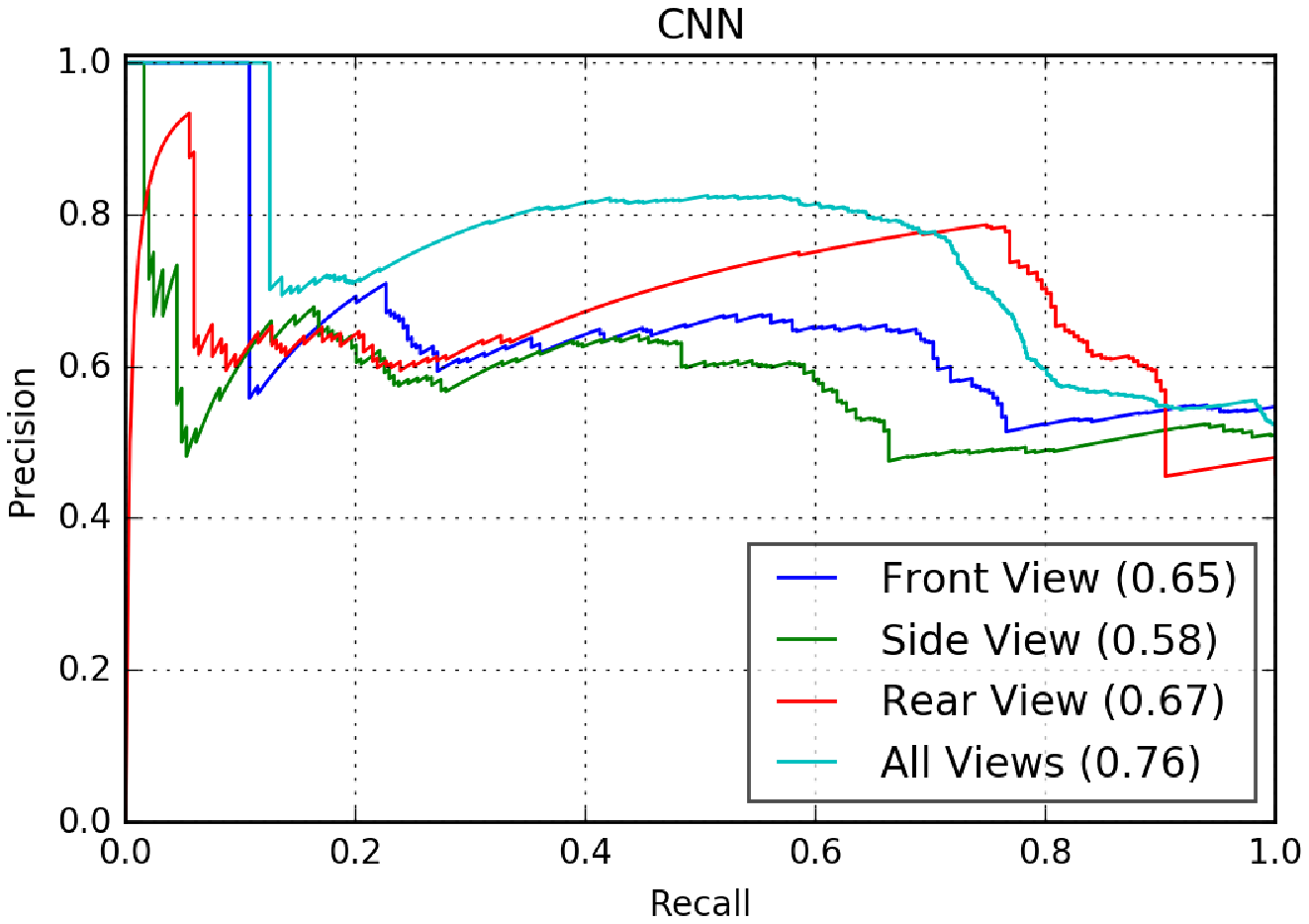}
    \hspace{1mm}
    \includegraphics[width=0.33\textwidth]{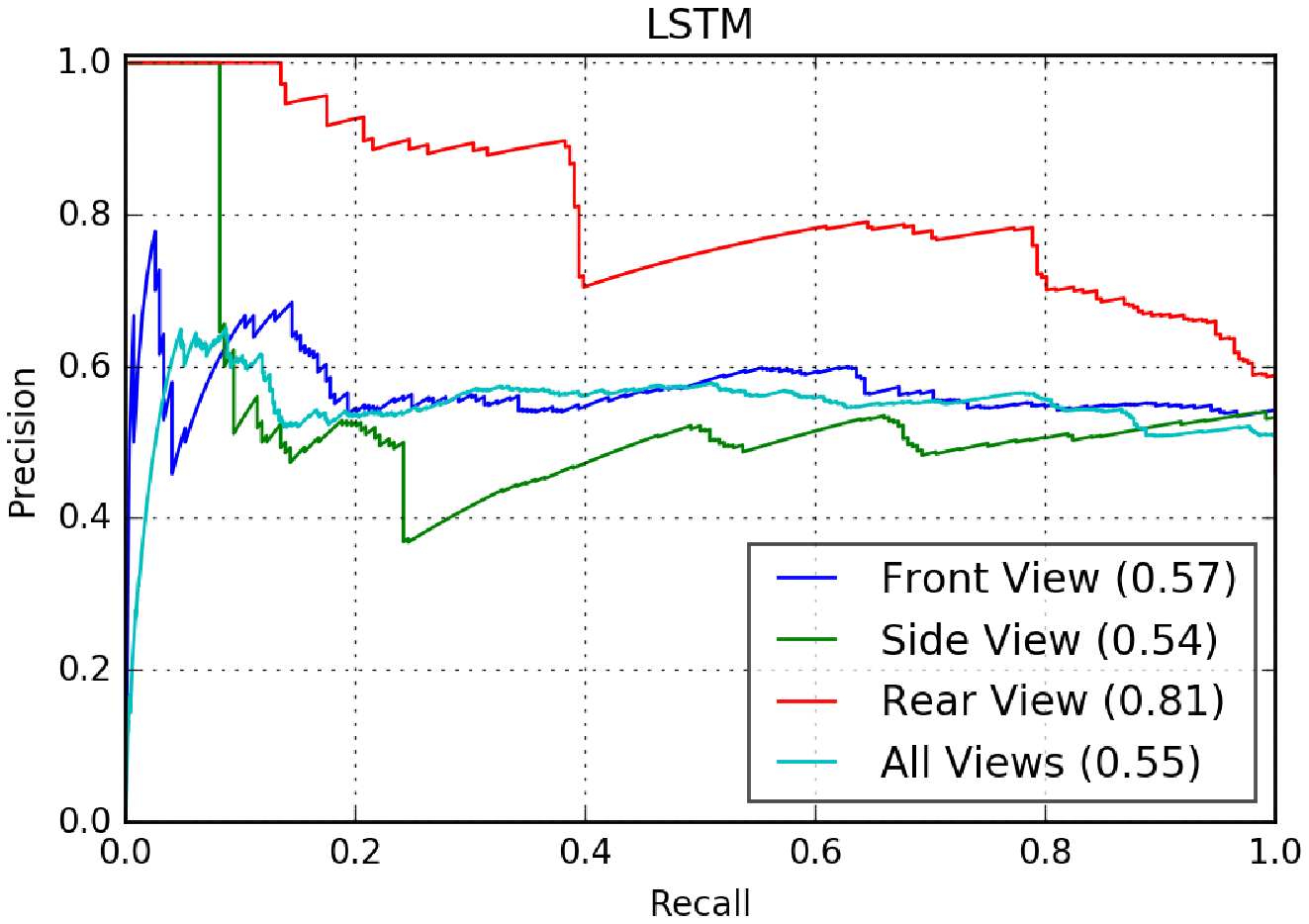}
    }
    \caption{Temporal classification performances using SVM, 1D CNN and LSTM. Average precision values are shown inside parenthesis. The curve `All Views' is obtained by a single classifier trained on data from all three views.}
    \label{fig:temporal-cls}
\end{figure}

\noindent
\textbf{Cross Validation} (CV) is a widely used model selection and evaluation method, particularly when the dataset is small~\cite{cv-survey-2010}. Leave-p-out and k-fold cross validation are commonly used CV methods. Leave-one-out (LOO) CV is a special case of leave-p-out CV with $p=1$. Each data point is left out for validation and the remaining points are used for training, and this is repeated for all data points in the dataset. It is exhaustive and considers all possible ways to divide the dataset into a training and validation sets.

Similar to leave-one-out CV, we employed a leave-one-car-out CV strategy, in which all data for a specific car is left out for validation and the remaining data is used for training, and this is repeated for all the cars in the dataset. The motivation for this CV strategy is to measure the generalization capability of our models across different car types. This is important, as it is not possible to collect training data for all types of cars in practice. Hence, the actual system performance will be higher than our measured performance, since the training set will contain the most frequent car makes in practice.

In training the neural networks, we used leave-two-cars-out CV with two validation sets $V_1$, $V_2$; one car's data for validation ($V_1$), one car's data ($V_2$, which is randomly selected) to decide how long to train the network to avoid overfitting (early stopping). We repeated this for all the $8$ cars and combined the results to estimate the classification accuracy. We also used two random re-starts for each CV round and took the network with the best validation performance (on the second validation set $V_2$).

We trained a separate classifier for each view (front, side, rear) and a single classifier for all the views.
We used cross validation as described above to evaluate the performance.
Figure~\ref{fig:temporal-cls} shows the precision-recall graphs and average precision (AP) values inside parenthesis. The results are with the manually annotated bounding boxes.
As expected, rear view idling classification performance is the highest and CNN and LSTM performed better than SVM due to their capability to model sequence data.
Front and side view performances are quite low, which is also expected, since maximum temperature profiles of idling and stopped cars are quite similar, as shown in Figure~\ref{fig:ir-temp-graphs}.
We need more complex features to model the spatio-temporal change of temperature for front and side views; this is described in the next section.

\section{Spatio-Temporal Modeling for Idling Classification}
\label{sec:sp-temp-cls}

In this section, we investigate spatio-temporal modeling of car temperature for idling classification to utilize the spatial distribution of car temperature as well as its temporal evolution. This can be achieved with sequence modeling approaches that use spatial features as input, e.g., recurrent neural networks (RNNs) that use convolutional features extracted from the individual images~\cite{video-deep-2018}, similar to~\cite{lrcnn-2015,lipreading-2016,tslstm-2017}. An alternative and more efficient way is to use CNNs, as in~\cite{rc3d,tcnn-2017}, with a stack of input frames sampled over the time interval. CNNs are easier to train and faster at test time compared to recurrent networks. 3D CNNs with 3D convolutions contain more parameters than 2D CNNs with 2D convolutions, and hence require much more training data and are not suitable for small datasets. Based on these insights, we experimented with two types of deep networks for classifying the spatio-temporal IR data: (1) 2D convolutional network (Section~\ref{sec:cnn}), (2) recurrent network with 2D convolutional feature extractor (Section~\ref{sec:cnn-rnn}). Both networks accept a 3D spatio-temporal volume of IR images as input.

The input to the networks has size $W \times H \times N$, where $W,H$ are the width and height of the car bounding boxes, N is the number of frames uniformly sampled over 3-minute (36 frames) IR image sequences. Since the recorded sequences are longer than 3 minutes, we slide a temporal window of stride 1 frame over the sequence and use each subsequence as a separate input sample, to increase the number of training/test samples. In the experiments, we sampled $N=7$ frames uniformly, i.e., one frame every $30$ seconds, and used a bounding box size of $W \times H = 100 \times 100$ pixels, since this was the typical size in the dataset. First a square size car bounding box is cropped from the IR frame, and then resized to $100 \times 100$. For side view, this square size box is cropped from the front part of the car, since the car's engine is there. The orientation of the car in side view (facing left or right) is determined simply by comparing the average temperatures on both ends; the front part has a higher temperature.

Deep networks are notorious for overfitting, especially on small datasets. To mitigate the overfitting problem, we used dropout regularization, as well as aggressive data augmentation as follows:
\begin{itemize}
 \item Random horizontal flip, with probability $0.5$
 \item Random image rotation with a maximum of $5$ degrees, with probability $0.5$
 \item Random erase or blur over random bounding boxes of maximum size  $10\times10$, with probability $0.5$
 \item Random blur with a maximum $\sigma=1.0$, and probability $0.5$
\end{itemize}

\subsection{Convolutional Network}
\label{sec:cnn}

\begin{figure}[ht]    
    \centerline{\includegraphics[width=0.8\textwidth]{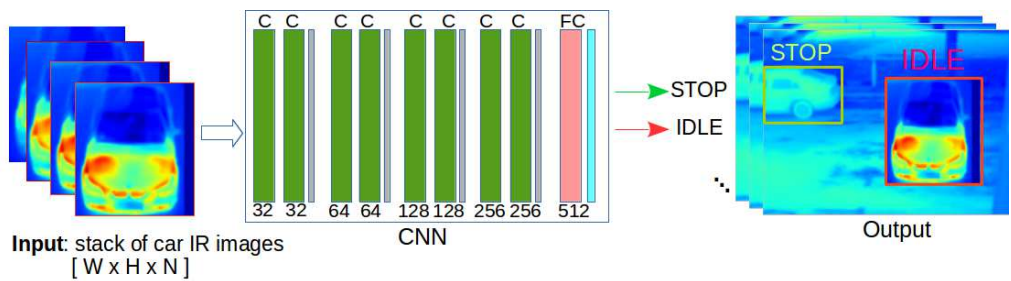}}
    \caption{Convolutional neural network for idling classification. The input to the network is a batch of 3D stack of $W \times H \times N$ cropped car bounding box IR images.}
    \label{fig:cnn-cls}
\end{figure}

We designed a 2D convolutional network with a 3D input tensor to model the spatio-temporal change of the car temperature, as shown in Figure~\ref{fig:cnn-cls}.
The input to the network is a tensor of size $B \times W \times H \times N$, where $B$ is the batch size.
The network architecture is 32$C(3,3)$, 32$C(3,3)$, MaxPool(2,2), 64$C(3,3)$, 64$C(3,3)$, MaxPool(2,2), Dropout(0.5), 128$C(3,3)$, 128$C(3,3)$, MaxPool(2,2), Dropout(0.5), 256$C(3,3)$, 256$C(3,3)$, MaxPool(2,2), Dropout(0.5), FC(512), Dropout(0.5), FC(2), Softmax. We manually explored smaller and larger architectures with different optimization algorithms and other hyperparameters and found this architecture to work the best.

We trained a single convolutional network for all the views.
We used Momentum optimizer with Nesterov momentum~\cite{nesterov-2013}, learning rate $0.002$, momentum $0.1$, exponential learning rate decay $0.96$ at $100$ steps to minimize the categorical cross entropy loss. The weights were initialized by the Xavier method~\cite{xavier}. The network was trained for a maximum of $100$ epochs.
We employed the same cross validation approach as described above (leave-two-cars-out CV) for model selection and performance evaluation.
We used TFLearn (TensorFlow)~\cite{tflearn} for implementation.
We observed that Adam optimizer leads to overfitting very quickly in a few epochs, in spite of regularization and data augmentation; Momentum optimizer turned out to be better in this respect, although it was much slower in convergence. This phenomenon has also been recently discovered by \cite{adam-sgd-arxiv17}.

\begin{figure}[ht]    
    \centerline{\includegraphics[width=0.5\textwidth]{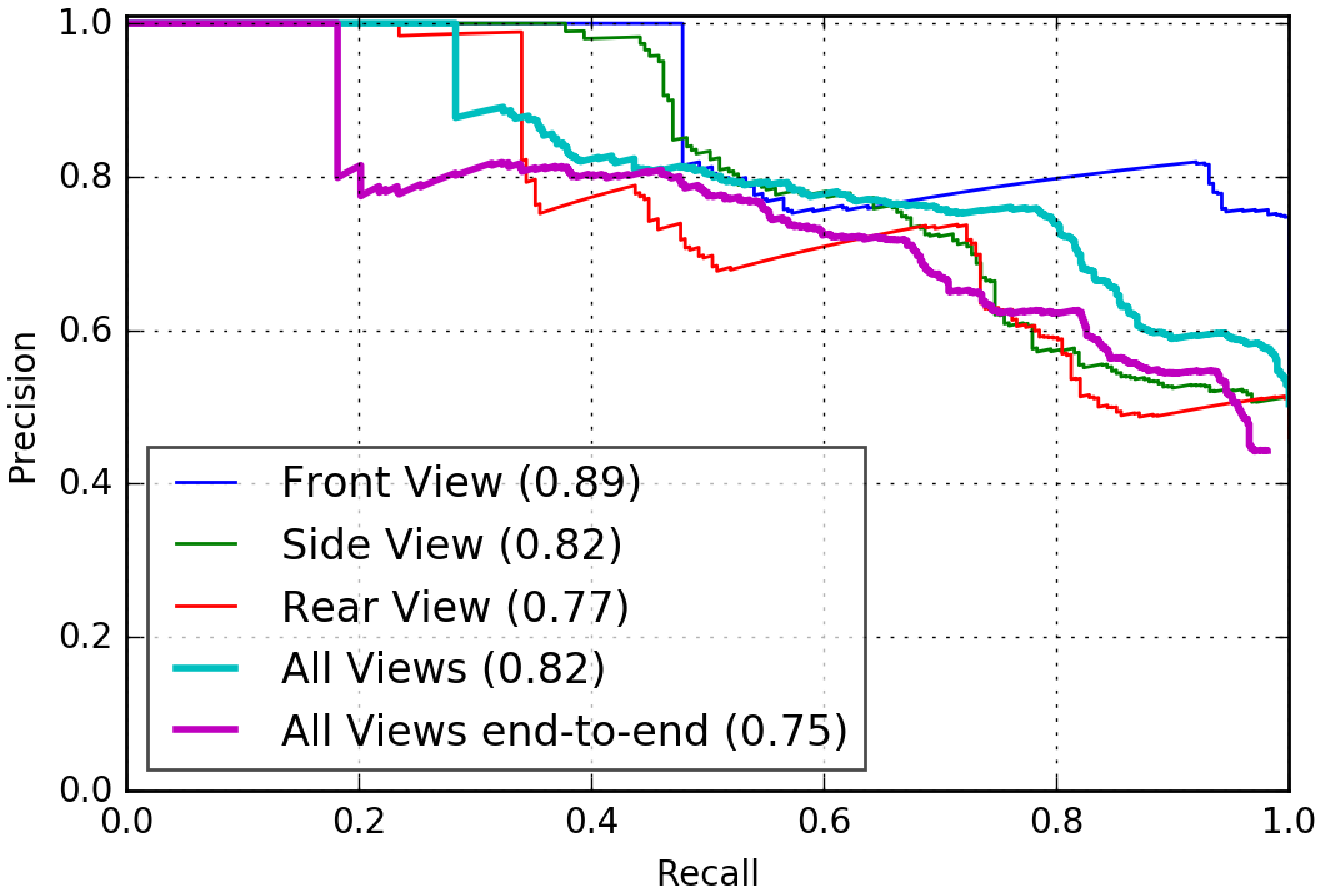}
                \includegraphics[width=0.5\textwidth]{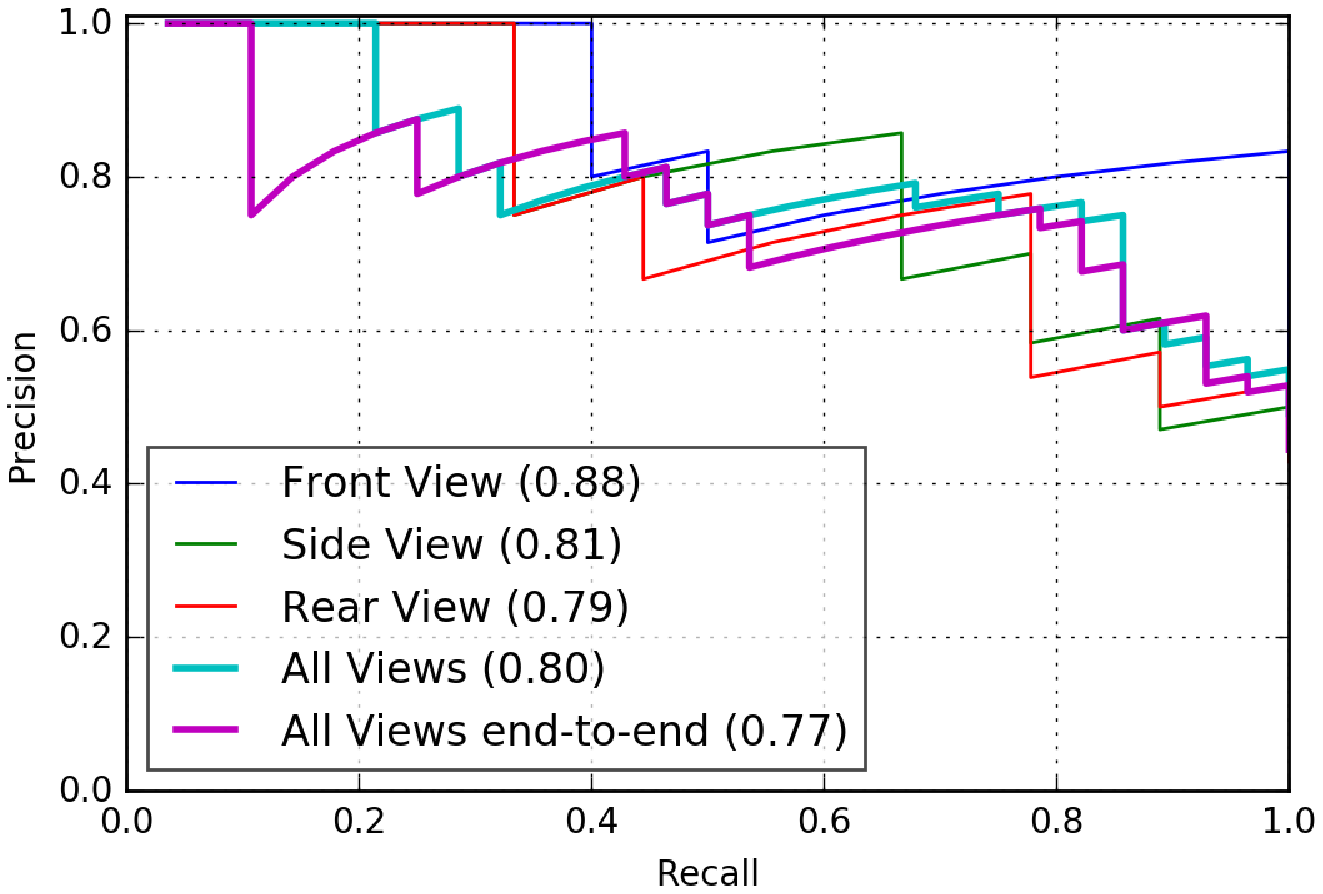}}
    \caption{Idling classification performance using 2D convolutional neural network and with 8-fold (leave-two-cars-out) cross validation.
    Left: subsequence-based evaluation. Right: sequence-based evaluation. Average precision values are shown inside parenthesis.}
    \label{fig:cls-cnn-pr}
\end{figure}

Figure~\ref{fig:cls-cnn-pr} shows idling classification performances using a single CNN for all the views. The view-based performances are also evaluated using the same network. We did two types of evaluations: (1) In sub-sequence-based evaluation, we consider each 3-minute subsequence as a test sample and evaluate accordingly, by comparing to the ground truth. (2) In sequence-based evaluation, we consider the whole sequence, e.g., 5-minutes. The matching is based on 50\% area, 90\% time overlap criteria in both cases. The score of a test sequence is computed as the average score of all its subsequences.

The end-to-end system (car detection and idling classification) performance is shown on the last line in each graph, with label `All Views end-to-end', which uses the manually annotated car bounding boxes in training and the output of car detector at evaluation/test. The other curves use manually labeled bounding boxes in both training and evaluation/test. This is to see the performance loss due to the errors in the car detector. The performance loss ($3-7$ points in average precision) in the end-to-end system is due to car detection and localization errors.

Spatio-temporal classification performance is significantly better than the temporal classification performance (Figure~\ref{fig:temporal-cls}), except the rear view classification performance, which is slightly higher in the temporal LSTM.
Among the 3 views, the front view performance is the highest. Even though the rear view seemed easier judging by the maximum temporal temperature profiles (Figure~\ref{fig:ir-image-view}), the CNN seems not to have learned it well, maybe because of the small area of the exhaust pipe compared to the whole car area. Another reason is that the exhaust pipe of one of the cars in the dataset was hidden and not visible in the IR images.

Based on these cross validation results, we can say that, the spatio-temporal CNN model is able to generalize across the majority of the car models, i.e., the CNN can learn spatio-temporal features to differentiate idling and stopped cars even if the cars are different and their spatio-temporal heat maps do not look visually very similar to the human eye (Figure~\ref{fig:ir-image-view}).
This means, we do not need to collect training data for all types of different car models, which would not be practical.
We also trained separate CNNs for each view, however, the performance was either the same or slightly lower. This might be due to the reduced amount of training data.
We also experimented with different sampling rates, N. For lower values of N, e.g., 3, 2, the performance drops; while there is no improvement for larger values, e.g., 9, 12.

The major difficulty is training and evaluating the CNN on a fairly small dataset. This results in overfitting in spite of aggressive regularization and data augmentation. Another problem is the difference between the training and validation data distributions, when the left-out car has a different spatio-temporal temperature signature from the cars in the training set and learned model performs poorly on the validation set. This is again due to the dataset size.

\subsection{Recurrent Network with Convolutional Features}
\label{sec:cnn-rnn}

\begin{figure}[ht]    
    \centerline{\includegraphics[width=0.8\textwidth]{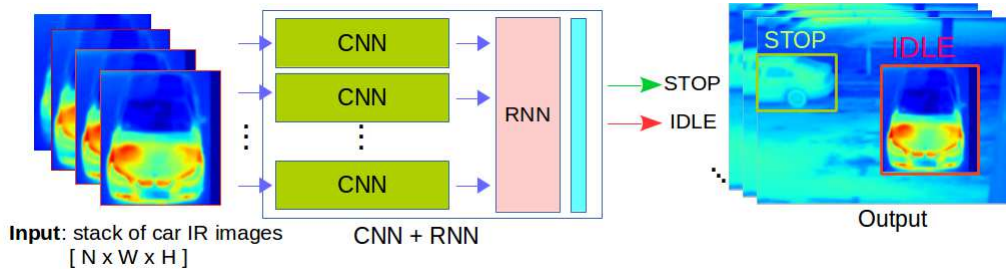}}
    \caption{Recurrent network with convolutional features for idling classification. Each $W \times H$ car bounding box image goes through the 2D CNN to extract features, which are fed to the RNN (LSTM) with $N=7$ time steps for sequence modeling.}
    \label{fig:cnn+rnn-cls}
\end{figure}

We also experimented with a recurrent network with a convolutional feature extractor, as shown in Figure~\ref{fig:cnn+rnn-cls}. Each cropped $W \times H$ car bounding box IR image goes through a convolutional feature extractor, and the output is fed to a recurrent neural network (LSTM) with $N=7$ time steps. $N$ frames are uniformly sampled per every 30 seconds as before. We trained a single network for all the views.

The architecture of the convolutional feature extractor is 32$C(3,3)$, 32$C(3,3)$, MaxPool(2,2), 64$C(3,3)$, 64$C(3,3)$, MaxPool(2,2), Droput(0.5), 80$C(3,3)$, 80$C(3,3)$, MaxPool(2,2), Droput(0.5), 96$C(3,3)$, 96$C(3,3)$, MaxPool(2,2), Droput(0.5). The output features are fed to LSTM(256, Dropout=0.5, Recurrent Droput=0.5), Dropout(0.5), FC(2), Softmax. The categorical cross entropy loss is optimized by Adam optimizer with learning rate $0.0001$. The weights are initialized by the Xavier method~\cite{xavier}. The network is trained for a maximum of $70$ epochs. We manually explored smaller and larger architectures with different optimization algorithms.

\begin{figure}[ht]    
    \centerline{\includegraphics[width=0.5\textwidth]{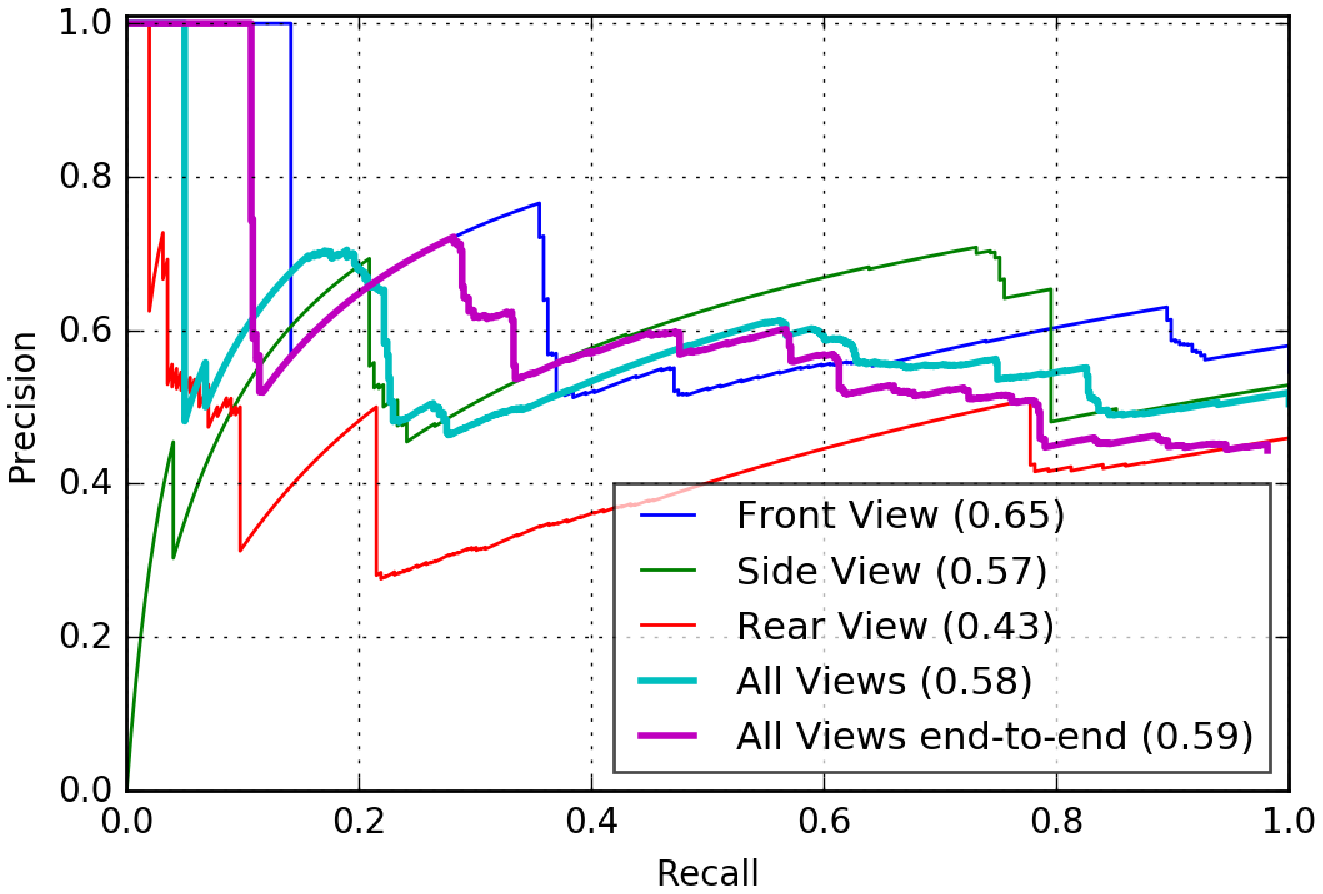}
                \includegraphics[width=0.5\textwidth]{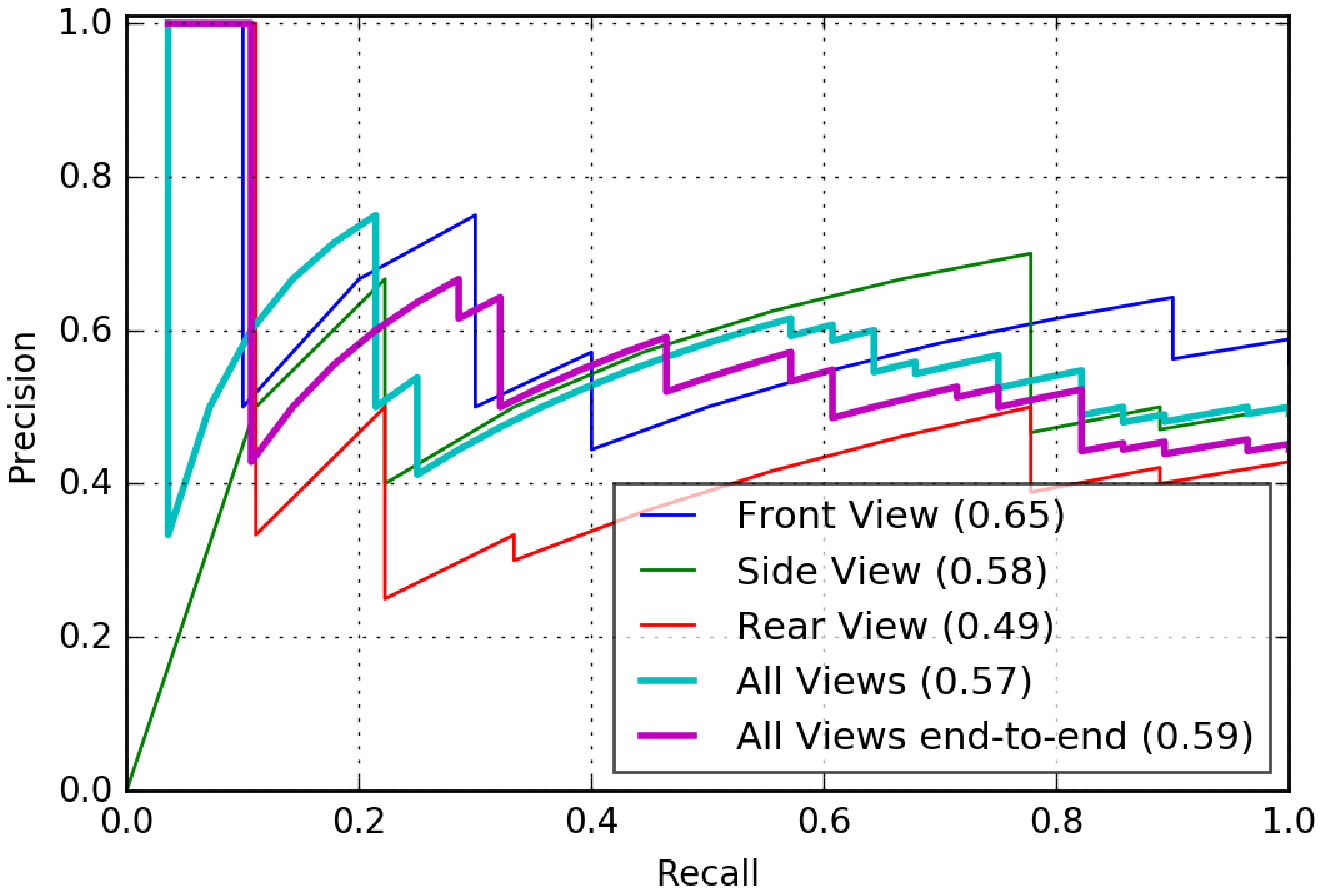}}
    \caption{Idling classification performance using a recurrent neural network with convolutional feature extractor.
    Left: subsequence-based evaluation. Right: sequence-based evaluation. Average precision values are shown inside parenthesis.}
    \label{fig:cls-rnn-pr}
\end{figure}

We used the same leave-two-cars-out cross validation as before to select the best model. Figure~\ref{fig:cls-rnn-pr} shows the idling classification performances for each view and for all the views, using the same network. The accuracy is significantly lower than that of CNN. The major problems are again (1) overfitting, in spite of aggressive regularization and data augmentation, (2) the difference between the training and validation data distributions, both due to the dataset size. We expect to obtain much higher accuracy with larger datasets. It is clear that the 2D CNN architecture with 3D inputs is more efficient and easier to train than the recurrent architecture. However, it is not clear which architecture would work better if the dataset was sufficiently large, or whether there are some architectures that would work much better with a suitable set of hyperparameters. On the current dataset and with the given architectures, the CNN is the winner.

\section{Conclusion}
We presented a novel vision application with promising early results: idling car detection with a thermal infrared camera. 
This is the first work to collect a IR car image sequence dataset and perform car detection and idling classification and evaluate various approaches.
We experimented with temporal and spatio-temporal modeling approaches using convolutional and recurrent networks. The convolutional networks worked better with around $0.80$ average precision values.
There is still ample room for performance improvement, especially if more data is available. 

A promising direction would be to train a Faster R-CNN-like network (or any object detection network) on 3D IR image volumes to directly detect the idling cars, instead of first detecting the cars and then performing idling classification. In this case, the network will learn the localization and classification simultaneously. The classifier layers can be either convolutional or recurrent. However, this will require abundant training data to work well. Finally, the same framework can be used to solve similar problems which require the detection of events in IR image sequences based on spatio-temporal temperature changes.

\section*{Acknowledgment}

This research was conducted as part of a joint research project with the National Environmental Agency (NEA) of Singapore, sponsored by the School of Electrical and Electronic Engineering, Nanyang Technological University, Singapore; EEE Seed Grant for Smart Nation Project, M4081921.040.

\bibliographystyle{spmpsci}
\bibliography{References}

\end{document}